\documentclass[twocolumn,nofootinbib,floatfix, superscriptaddress]{revtex4} %

\usepackage{dcolumn}

\usepackage{times}

\usepackage{amsmath}
\usepackage{amsfonts}
\usepackage{amssymb}
\usepackage{graphicx}
\newcommand{\maxpr}{\overline{p}_r}
\newcommand{\like}{\mathcal{L}}
\newcommand{\remove}[1]{}

\begin{document}

\title{Hierarchical structure and the prediction of missing links in networks\footnote{This paper was published as {\em Nature} {\bf 453}, 98 -- 101 (2008); \\ doi:10.1038/nature06830.}}

\author{Aaron Clauset}
\affiliation{Department of Computer Science, University of New Mexico, Albuquerque, NM 87131, USA}
\affiliation{Santa Fe Institute, 1399 Hyde Park Road, Santa Fe NM, 87501, USA}
\author{Cristopher Moore}
\affiliation{Department of Computer Science, University of New Mexico, Albuquerque, NM 87131, USA}
\affiliation{Santa Fe Institute, 1399 Hyde Park Road, Santa Fe NM, 87501, USA}
\affiliation{Department of Physics and Astronomy, University of New Mexico, Albuquerque, NM 87131, USA}
\author{M. E. J. Newman}
\affiliation{Santa Fe Institute, 1399 Hyde Park Road, Santa Fe NM, 87501, USA}
\affiliation{Department of Physics and Center for the Study of Complex Systems, University of Michigan, Ann Arbor, MI 48109, USA}

\maketitle 

\textbf{
Networks have in recent years emerged as an invaluable tool for
 describing and quantifying complex systems in many branches of
 science~\cite{wasserman:faust:1994,albert:barabasi:2002,newman:2003}.
 Recent studies suggest that networks often exhibit hierarchical
 organization, where vertices divide into groups that further subdivide
 into groups of groups, and so forth over multiple scales.  In many cases
 these groups are found to correspond to known functional units, such as
 ecological niches in food webs, modules in biochemical networks (protein
 interaction networks, metabolic networks, or genetic regulatory networks), 
 or communities in social networks~\cite{ravasz:somera:mongru:oltavai:barabasi:2002,clauset:newman:moore:2004,guimera:amaral:2005,lagomarsino:etal:2007}.
 Here we present a general technique for inferring hierarchical structure
 from network data and demonstrate that the existence of hierarchy can
 simultaneously explain and quantitatively reproduce many commonly
 observed topological properties of networks, such as right-skewed degree
 distributions, high clustering coefficients, and short path lengths.  We
 further show that knowledge of hierarchical structure can be used to
 predict missing connections in partially known networks with high
 accuracy, and for more general network structures than competing
 techniques~\cite{liben-nowell:kleinberg:2003}.  Taken together, our
 results suggest that hierarchy is a central organizing principle of
 complex networks, capable of offering insight into many network
 phenomena.
}

\begin{figure}[!b]
\begin{center}
\includegraphics[scale=0.3]{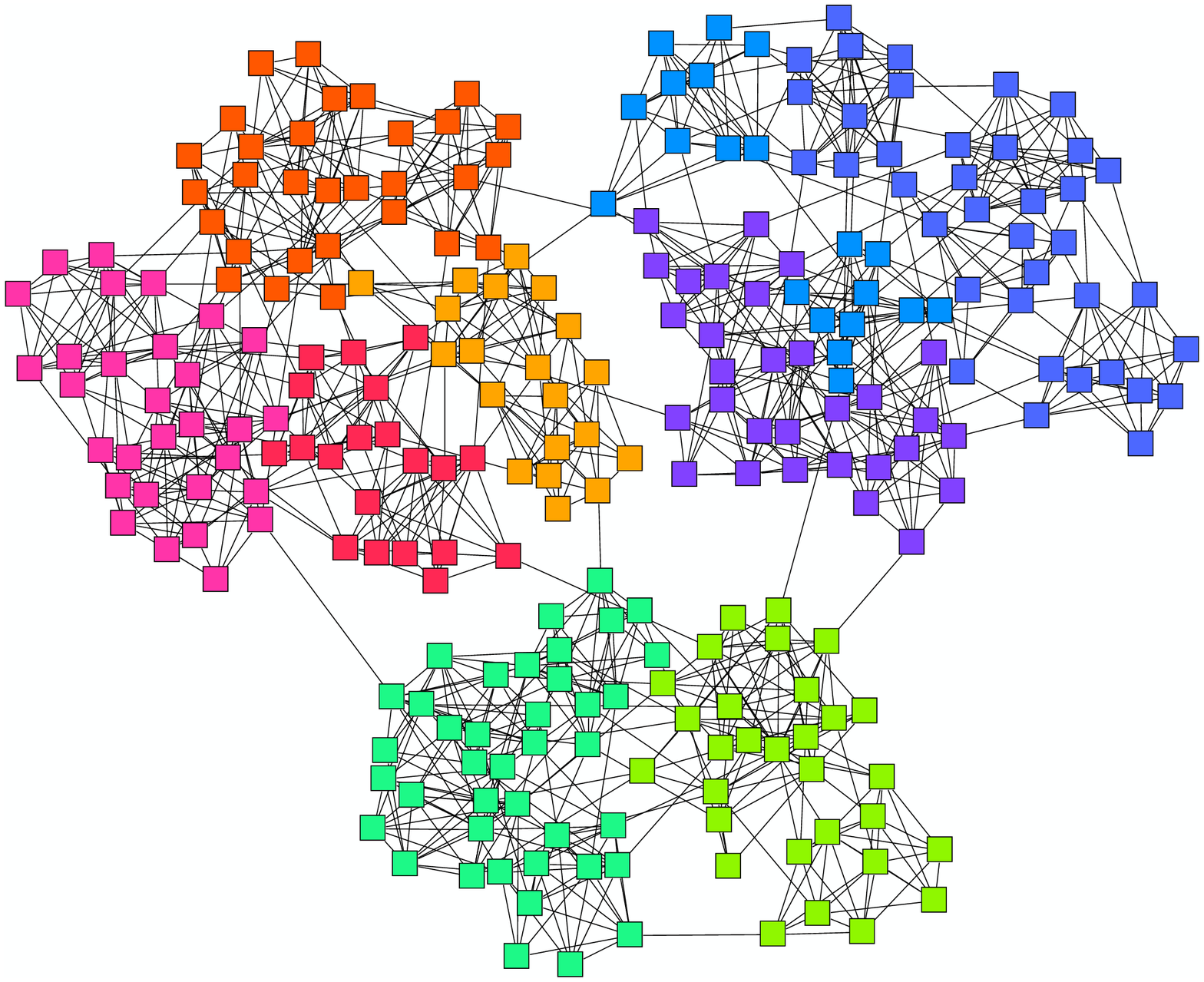} \\
\includegraphics[scale=0.4]{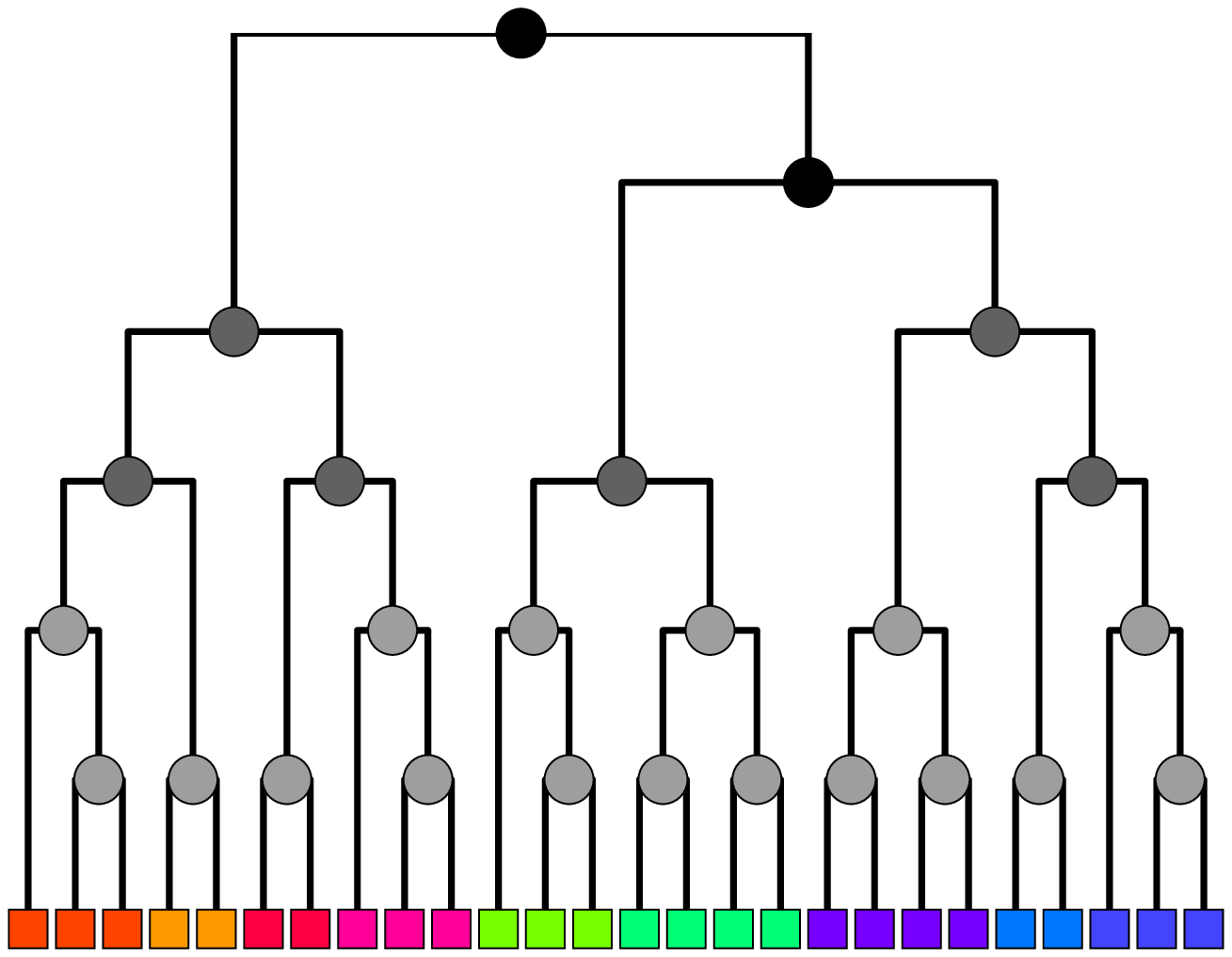}
\caption{A hierarchical network with structure on many scales and the corresponding hierarchical random graph.  Each internal node~$r$ of the dendrogram is associated with a probability~$p_{r}$ that a pair of vertices in the left and right subtrees of that node are connected.  (The shades of the internal nodes in the figure represent the probabilities.) }
\label{fig:example}
\end{center}
\end{figure}

A great deal of recent work has been devoted to the study of clustering and community structure in networks~\cite{girvan:newman:2002,krause:frank:mason:ulanowicz:taylor:2003,radicchi:castellano:cecconi:loreto:parisi:2004,clauset:newman:moore:2004,guimera:amaral:2005}. Hierarchical structure goes beyond simple clustering, however, by explicitly including organization at all scales in a network simultaneously.  Conventionally, hierarchical structure is represented by a tree or \emph{dendrogram} in which closely related pairs of vertices have lowest common ancestors that are lower in the tree than those of more distantly related pairs---see Fig.~\ref{fig:example}.  We expect the probability of a connection between two vertices to depend on their degree of relatedness.  Structure of this type can be modelled mathematically using a probabilistic approach in which we endow each internal node~$r$ of the dendrogram with a probability~$p_{r}$ and then connect each pair of vertices for whom~$r$ is the lowest common ancestor independently with probability~$p_{r}$ (Fig.~\ref{fig:example}).

This model, which we call a \emph{hierarchical random graph}, is similar in spirit (although different in realization) to the tree-based models used in some studies of network search and navigation~\cite{WDN02,Kleinberg02}. Like most work on community structure, it assumes that communities at each level of organization are disjoint.  Overlapping communities have occasionally been studied (see, for example~\cite{vicsek}) and could be represented using a more elaborate probabilistic model, but as we discuss below the present model already captures many of the structural features of interest.

Given a dendrogram and a set of probabilities~$p_{r}$, the hierarchical random graph model allows us to generate artificial networks with a specified hierarchical structure, a procedure that might be useful in certain situations.  Our goal here, however, is a different one.  We would like to detect and analyze the hierarchical structure, if any, of networks in the real world.  We accomplish this by fitting the hierarchical model to observed network data using the tools of statistical inference, combining a maximum likelihood approach~\cite{casella:berger:1990} with a Monte Carlo sampling algorithm~\cite{newman:barkema:1999} on the space of all possible dendrograms.  This technique allows us to sample hierarchical random graphs with probability proportional to the likelihood that they generate the observed network.  To obtain the results described below we combine information from a large number of such samples, each of which is a reasonably likely model of the data.

The success of this approach relies on the flexible nature of our hierarchical model, which allows us to fit a wide range of network structures.  The traditional picture of communities or modules in a network, for example, corresponds to connections that are dense within groups of vertices and sparse between them---a behaviour called ``assortativity'' in the literature~\cite{newman:2002}.  The hierarchical random graph can capture behaviour of this kind using probabilities~$p_{r}$ that decrease as we move higher up the tree.  Conversely, probabilities that increase as we move up the tree correspond to ``disassortative'' structures in which vertices are less likely to be connected on small scales than on large ones.  By letting the~$p_{r}$values vary arbitrarily throughout the dendrogram, the hierarchical random graph can capture both assortative and disassortative structure, as well as arbitrary mixtures of the two, at all scales and in all parts of the network.

To demonstrate our method we have used it to construct hierarchical decompositions of three example networks drawn from disparate fields: the metabolic network of the spirochete \emph{Treponema pallidum}~\cite{huss:holme:2007}, a network of associations between terrorists~\cite{krebs:2002}, and a food web of grassland species~\cite{dawah:hawkins:claridge:1995}.  To test whether these decompositions accurately capture the networks' important structural features, we use the sampled dendrograms to generate new networks, different in detail from the originals but, by definition, having similar hierarchical structure (see the Supplementary Information for more details).  We find that these ``resampled'' networks match the statistical properties of the originals closely, including their degree distributions, clustering coefficients, and distributions of shortest path lengths between pairs of vertices, despite the fact that none of these properties is explicitly represented in the hierarchical random graph (Table~\ref{table:resampled}, and Fig.~\ref{fig:valid} in the Supplementary Information). Thus it appears that a network's hierarchical structure is capable of explaining a wide variety of other network features as well.

\begin{table}
\begin{center}
\begin{tabular}{lc|cc|cc|cc} 
Network & & $\langle k \rangle_{\rm real}$ & $\langle k \rangle_{\rm samp}$
& $C_{\rm real}$ & $ C_{\rm samp}$ & $d_{\rm real}$ & $d_{\rm samp}$ \\
\hline
\emph{T. pallidum} & & 4.8 & 3.7(1) & 0.0625 & 0.0444(2) & 3.690 & 3.940(6) \\
Terrorists                  & & 4.9 & 5.1(2) & 0.361 & 0.352(1) & 2.575 & 2.794(7) \\ 
Grassland                & & 3.0 & 2.9(1) & 0.174 & 0.168(1) & 3.29 & 3.69(2) 
\end{tabular}
\caption{Comparison of network statistics for the three example networks studied and new networks generated by resampling from our hierarchical model.  The generated networks closely match the average degree~$\langle k \rangle$, clustering coefficient $C$, and average vertex-vertex distance d in each case, suggesting that they capture much of the real networks' structure.  Parenthetical values indicate standard errors on the final digits.}
\label{table:resampled}
\end{center}
\end{table}

\begin{figure}[t]
\begin{center}
\includegraphics[scale=0.38]{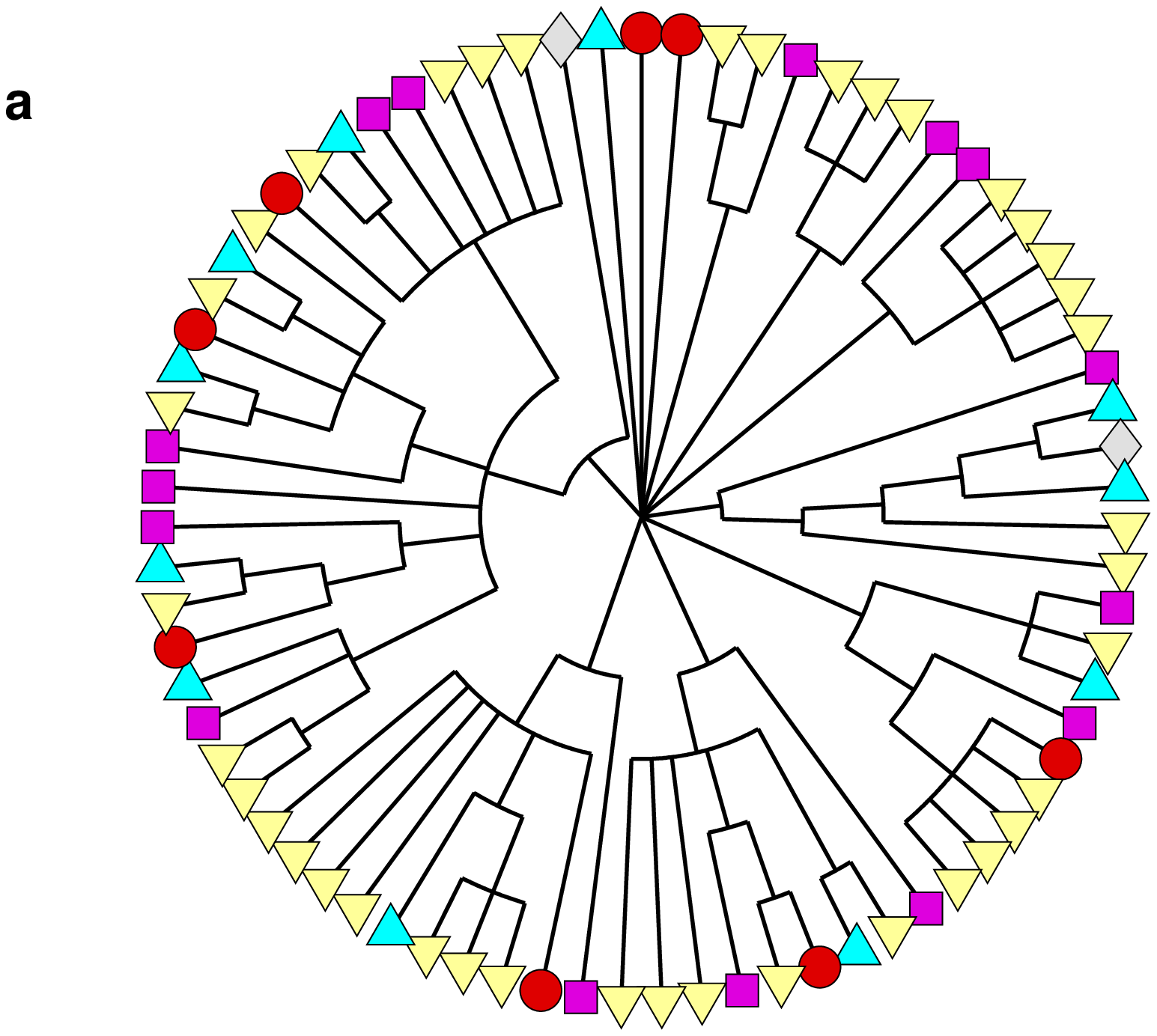} \\
\includegraphics[scale=0.31]{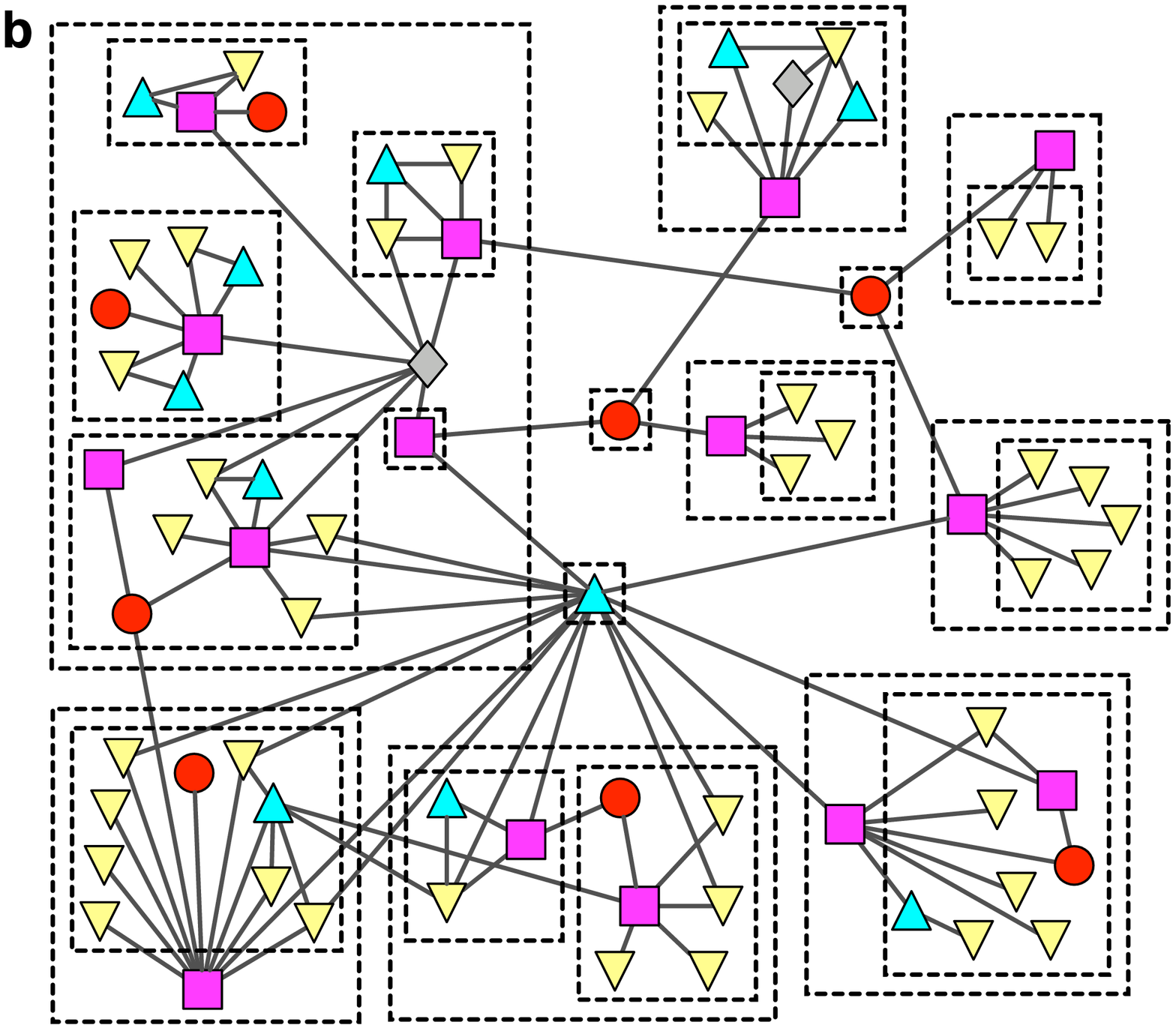}
\caption{Application of the hierarchical decomposition to the network of grassland species interactions. \textbf{a},~Consensus dendrogram reconstructed from the sampled hierarchical models. \textbf{b},~A visualization of the network in which the upper few levels of the consensus dendrogram are shown as boxes around species (plants, herbivores, parasitoids, hyper-parasitoids and hyper-hyper-parasitoids are shown as circles, boxes, down triangles, up triangles and diamonds respectively).  Note that in several cases, a set of parasitoids is grouped into a disassortative community by the algorithm, not because they prey on each other, but because they prey on the same herbivore.}
\label{fig:consensus}
\end{center}
\end{figure}

The dendrograms produced by our method are also of interest in themselves, as a graphical representation and summary of the hierarchical structure of the observed network.  As discussed above, our method can generates not just a single dendrogram but a set of dendrograms, each of which is a good fit to the data.  From this set we can, using techniques from phylogeny reconstruction~\cite{bryant:2003}, create a single \emph{consensus dendrogram}, which captures the topological features that appear consistently across all or a large fraction of the dendrograms and typically represents a better summary of the network's structure than any individual dendrogram. Figure~\ref{fig:consensus}a shows such a consensus dendrogram for the grassland species network, which clearly reveals communities and sub-communities of plants, herbivores, parasitoids, and hyper-parasitoids.

Another application of the hierarchical decomposition is the prediction of missing interactions in networks.  In many settings, the discovery of interactions in a network requires significant experimental effort in the laboratory or the field.  As a result, our current pictures of many networks are substantially incomplete~\cite{dunne:williams:martinez:2002,szilagyi:etal:2002,sprinzak:sattath:margalit:2003,ito:chiba:etal:2001,lakhina,clauset:moore:2005,martinez:etal:1999}. An attractive alternative to checking exhaustively for a connection between every pair of vertices in a network is to try to predict, in advance and based on the connections already observed, which vertices are most likely to be connected, so that scarce experimental resources can be focused on testing for those interactions.  If our predictions are good, we can in this way reduce substantially the effort required to establish the network's topology.

The hierarchical decomposition can be used as the basis for an effective method of predicting missing interactions as follows.  Given an observed but incomplete network, we generate as described above a set of hierarchical random graphs---dendrograms and the associated probabilities~$p_{r}$---that fit that network.  Then we look for pairs of vertices that have a high average probability of connection within these hierarchical random graphs but which are unconnected in the observed network.  These pairs we consider the most likely candidates for missing connections.  (Technical details of the procedure are given in the Supplementary Information.)

\begin{figure}[t]
\begin{center}
\includegraphics[scale=0.4]{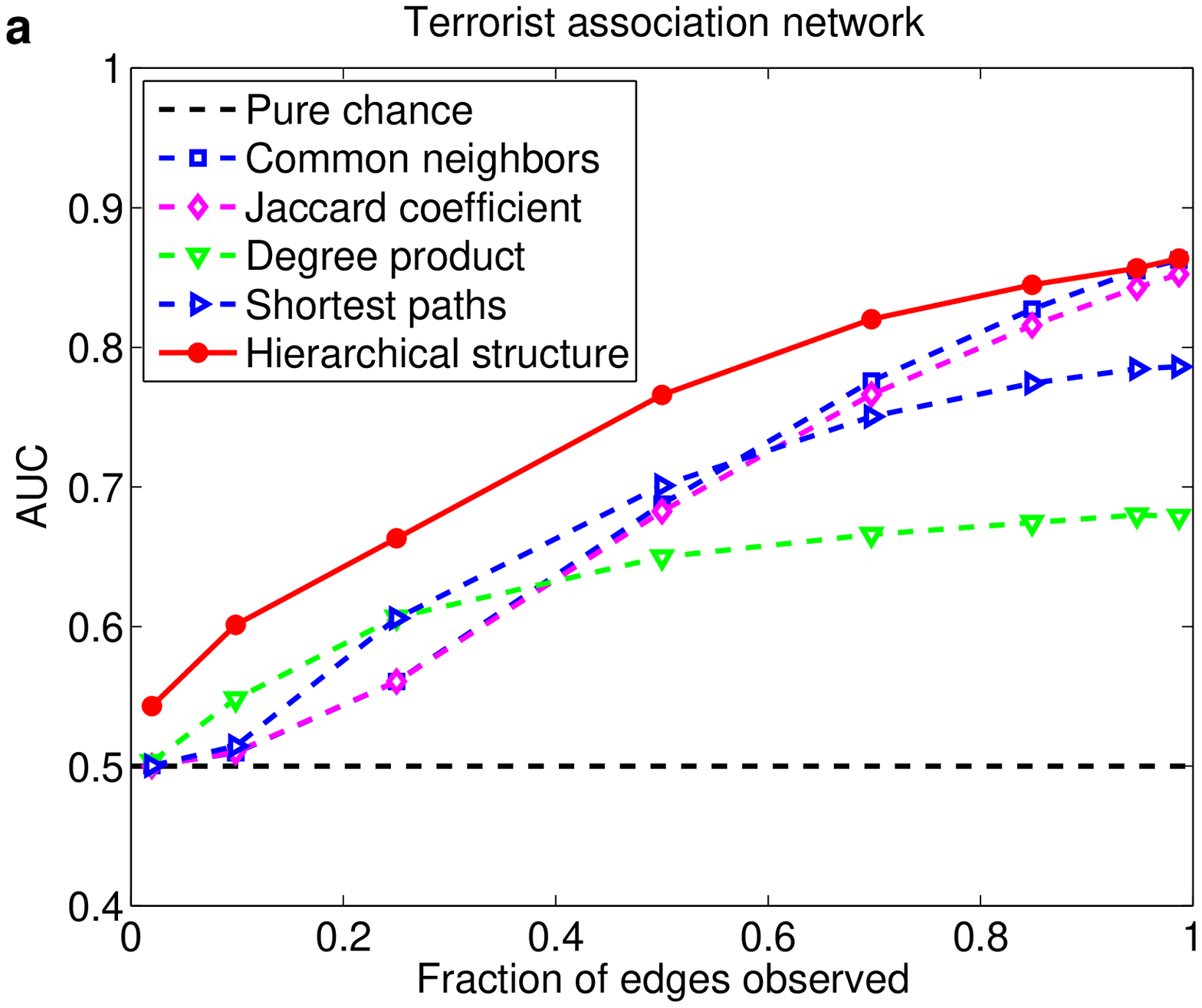} 
\includegraphics[scale=0.4]{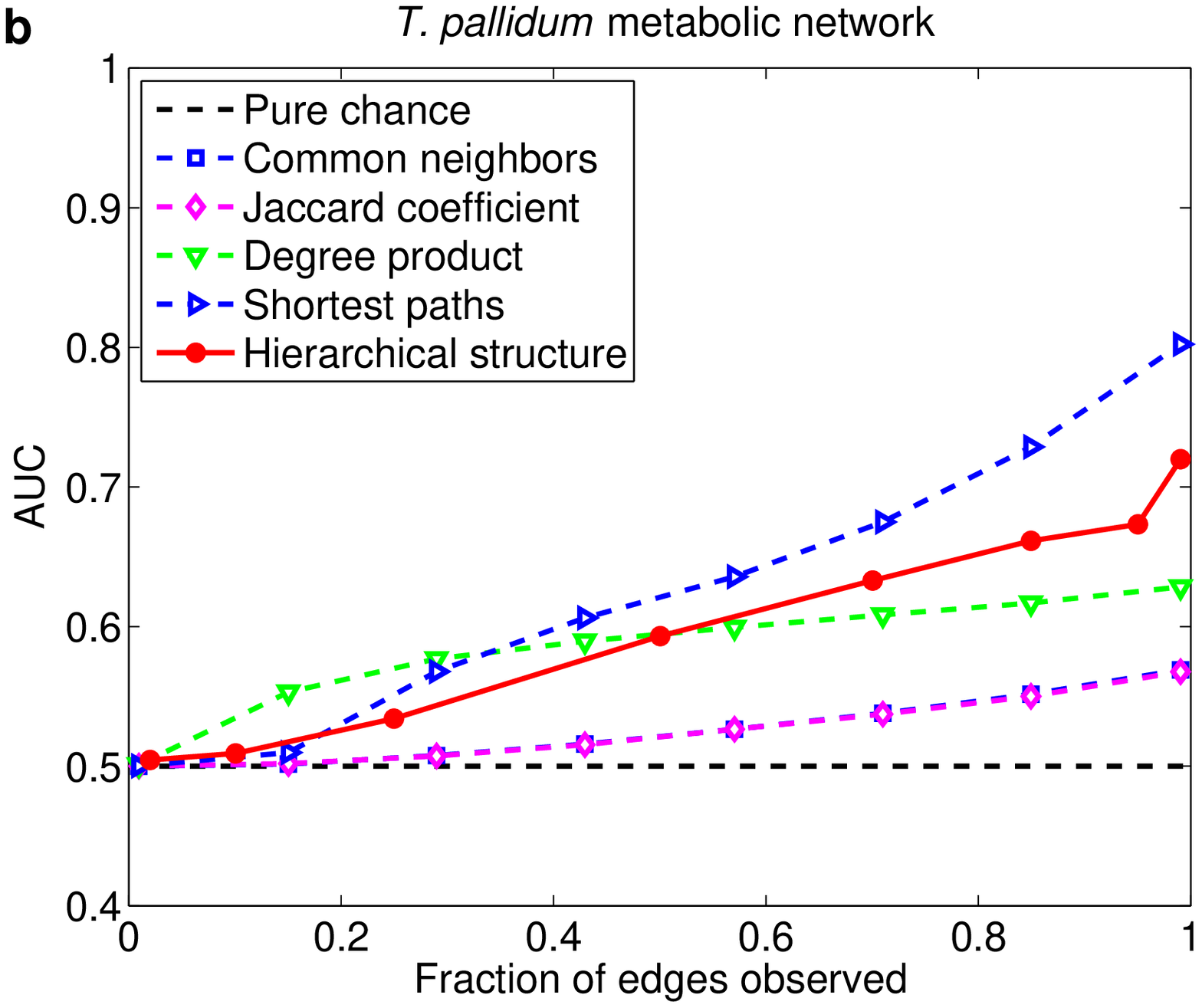} 
\includegraphics[scale=0.4]{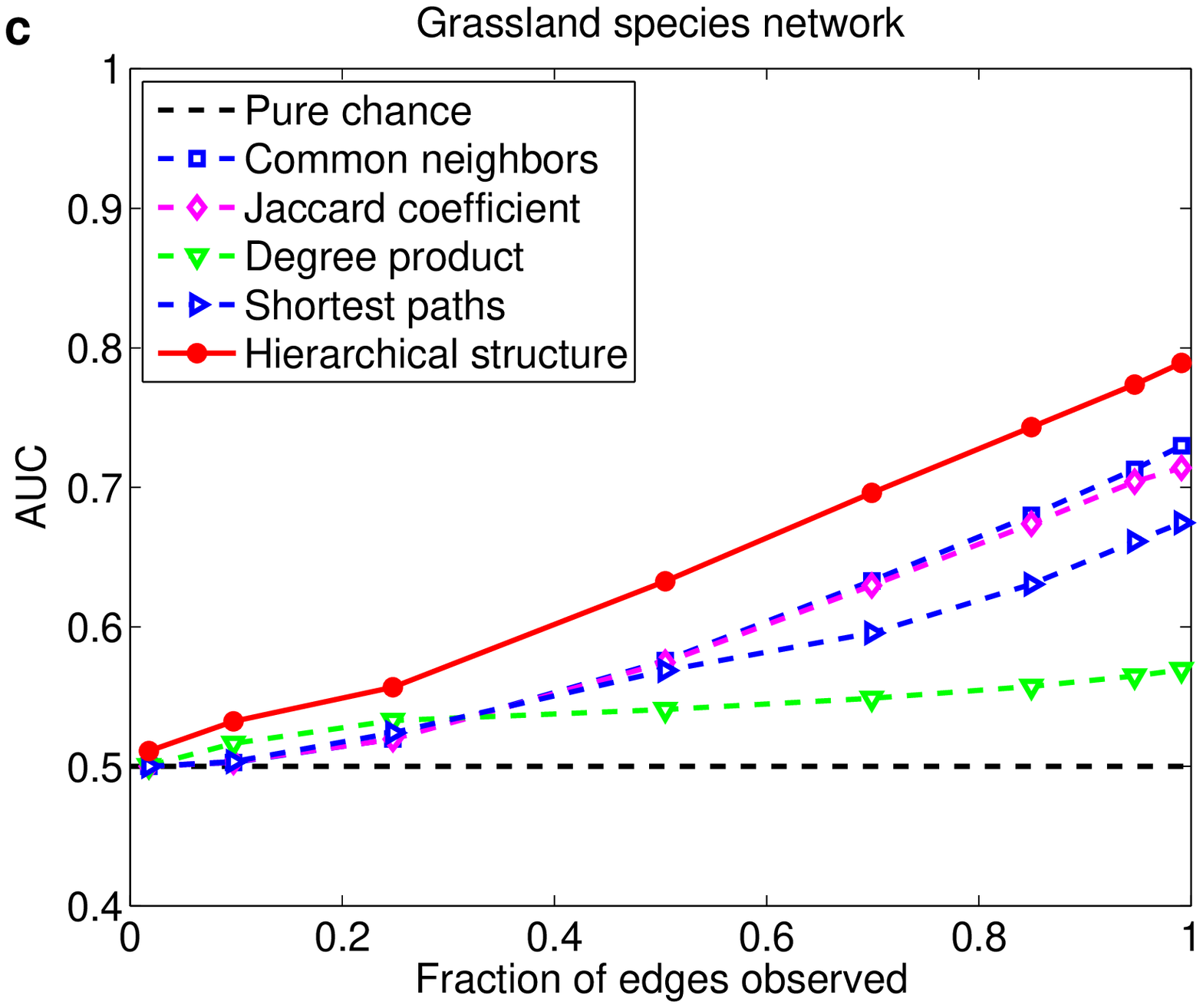} 
\caption{Comparison of link prediction methods. Average AUC statistic, i.e., the probability of ranking a true positive over a true negative, as a function of the fraction of connections known to the algorithm, for the link prediction method presented here and a variety of previously published methods.
}
\label{fig:predictions2}
\end{center}
\end{figure}

We demonstrate the method using our three example networks again.  For each network we remove a subset of connections chosen uniformly at random and then attempt to predict, based on the remaining connections, which ones have been removed.  A standard metric for quantifying the accuracy of prediction algorithms, commonly used in the medical and machine learning communities, is the AUC statistic, which is equivalent to the area under the receiver-operating characteristic (ROC) curve~\cite{hanley}.  In the present context, the AUC statistic can be interpreted as the probability that a randomly chosen missing connection (a true positive) is given a higher score by our method than a randomly chosen pair of unconnected vertices (a true negative).  Thus, the degree to which the AUC exceeds $1/2$ indicates how much better our predictions are than chance.  Figure~\ref{fig:predictions2} shows the AUC statistic for the three networks as a function of the fraction of the connections known to the algorithm.  For all three networks our algorithm does far better than chance, indicating that hierarchy is a strong general predictor of missing structure. It is also instructive to compare the performance of our method to that of other methods for link prediction~\cite{liben-nowell:kleinberg:2003}. Previously proposed methods include assuming that vertices are likely to be connected if they have many common neighbours, if there are short paths between them, or if the product of their degrees is large.  These approaches work well for strongly assortative networks such as the collaboration and citation networks~\cite{liben-nowell:kleinberg:2003} and for the metabolic and terrorist networks studied here (Fig.~\ref{fig:predictions2}a,b).  Indeed, for the metabolic network the shortest-path heuristic performs better than our algorithm.

However, these simple methods can be misleading for networks that exhibit more general types of structure.  In food webs, for instance, pairs of predators often share prey species, but rarely prey on each other. In such situations a common-neighbour or shortest-path-based method would predict connections between predators where none exist.  The hierarchical model, by contrast, is capable of expressing both assortative and disassortative structure and, as Fig.~\ref{fig:predictions2}c shows, gives substantially better predictions for the grassland network.  (Indeed, in Fig.~\ref{fig:consensus}b there are several groups of parasitoids that our algorithm has grouped together in a disassortative community, in which they prey on the same herbivore but not on each other.)  The hierarchical method thus makes accurate predictions for a wider range of network structures than the previous methods.

In the applications above, we have assumed for simplicity that there are no false positives in our network data, i.e., that every observed edge corresponds to a real interaction.  In networks where false positives may be present, however, they too could be predicted using the same approach: we would simply look for pairs of vertices that have a \emph{low} average probability of connection within the hierarchical random graph but which are connected in the observed network.

The method described here could also be extended to incorporate domain-specific information, such as speciesÕ morphological or behavioural traits for food webs~\cite{martinez:etal:1999} or phylogenetic or binding-domain data for biochemical networks~\cite{szilagyi:etal:2002}, by adjusting the probabilities of edges accordingly.  As the results above show, however, we can obtain good predictions even in the absence of such information, indicating that topology alone can provide rich insights.

In closing, we note that our approach differs crucially from previous work on hierarchical structure in networks~\cite{wasserman:faust:1994,ravasz:somera:mongru:oltavai:barabasi:2002,girvan:newman:2002,radicchi:castellano:cecconi:loreto:parisi:2004,clauset:newman:moore:2004,guimera:amaral:2005,salespardo:2007,lagomarsino:etal:2007} in that it acknowledges explicitly that most real-world networks have many plausible hierarchical representations of roughly equal likelihood. Previous work, by contrast, has typically sought a single hierarchical representation for a given network.  By sampling an ensemble of dendrograms, our approach avoids over-fitting the data and allows us to explain many common topological features, generate resampled networks with similar structure to the original, derive a clear and concise summary of a network's structure via its consensus dendrogram, and accurately predict missing connections in a wide variety of situations.

\vspace{2mm}
\noindent \textbf{Acknowledgments}: The authors thank Jennifer Dunne, Michael Gastner, Petter Holme, Mikael Huss, Mason Porter, Cosma Shalizi and Chris Wiggins for their help, and the Santa Fe Institute for its support. C.M. thanks the Center for the Study of Complex Systems at the University of Michigan for their hospitality while some of this work was conducted. Computer code implementing many of the analysis methods described in this paper can be found online at http://www.santafe.edu/\verb+~+aaronc/randomgraphs/\ .

\renewcommand{\thefigure}{S\arabic{figure}}
\setcounter{figure}{0}
\renewcommand{\thetable}{S\arabic{table}}
\setcounter{table}{0}

\newpage
\begin{appendix}
\section*{Supplementary Information}

\section{Hierarchical random graphs}

Our model for the hierarchical organization of a network is as 
follows.\footnote{Computer code implementing many of the analysis methods 
described in this paper can be found online at \\
{\tt www.santafe.edu/$\sim$aaronc/randomgraphs/}.}
Let~$G$ be a graph with~$n$ vertices.  A \emph{dendrogram}~$D$ is a binary
tree with~$n$ leaves corresponding to the vertices of~$G$.  Each of
the~$n-1$ internal nodes of~$D$ corresponds to the group of vertices that are
descended from it.  We associate a probability~$p_r$ with each internal
node~$r$.  Then, given two vertices~$i,j$ of~$G$, the probability~$p_{ij}$
that they are connected by an edge is~$p_{ij} = p_r$ where~$r$ is their
lowest common ancestor in~$D$.  The combination~$(D, \{p_r\})$ of the
dendrogram and the set of probabilities then defines a \emph{hierarchical
 random graph}.

Note that if a community has, say, three subcommunities, with an equal
probability $p$ of connections between them, we can represent this in our
model by first splitting one of these subcommunities off, and then
splitting the other two.  The two internal nodes corresponding to these
splits would be given the same probabilities $p_r=p$.  This yields three
possible binary dendrograms, which are all considered equally likely.

We can think of the hierarchical random graph as a variation on the
classical Erd\H{o}s--R\'enyi random graph~$G(n,p)$.  As in that model, the
presence or absence of an edge between any pair of vertices is independent
of the presence or absence of any other edge.  However, whereas in~$G(n,p)$
every pair of vertices has the same probability~$p$ of being connected, in
the hierarchical random graph the probabilities are inhomogeneous, with the
inhomogeneities controlled by the topological structure of the
dendrogram~$D$ and the parameters~$\{p_r\}$.  Many other models with
inhomogeneous edge probabilities have, of course, been studied in the past.
One example is 
a structured random graph in which there are a finite
number of types of vertices with a matrix $p_{kl}$ giving the connection
probabilities between them.\footnote{F. McSherry, ``Spectral Partitioning
 of Random Graphs.'' \textit{Proc. Foundations of Computer Science
   (FOCS)}, pp.  529--537 (2001)}

\section{Fitting the hierarchical random graph to data}

Now we turn to the question of finding the hierarchical random graph or
graphs that best fits the observed real-world network~$G$.  Assuming that
all hierarchical random graphs are \emph{a priori} equally likely, the
probability that a given model~$(D,\{p_r\})$ is the correct explanation of
the data is, by Bayes' theorem, proportional to the posterior probability
or \emph{likelihood}~$\like$ with which that model generates the observed
network.\footnote{G. Casella and R.~L. Berger, ``Statistical Inference.''
 Duxbury Press, Belmont (2001).}  Our goal is to maximize~$\like$ or, more
generally, to sample the space of all models with probability proportional 
to~$\like$.

Let~$E_r$ be the number of edges in~$G$ whose endpoints have~$r$ as their
lowest common ancestor in~$D$, and let~$L_r$ and~$R_r$, respectively, be
the numbers of leaves in the left and right subtrees rooted at~$r$.  Then
the likelihood of the hierarchical random graph 
is
\begin{align}
\like(D,\{p_r\}) = \prod_{r \in D} p_r^{E_r}
                  \left( 1 - p_r \right)^{L_rR_r-E_r}
\label{eq:like1}
\end{align}
with the convention that~$0^0=1$.

\begin{figure}[t]
\begin{center}
\includegraphics[scale=0.35]{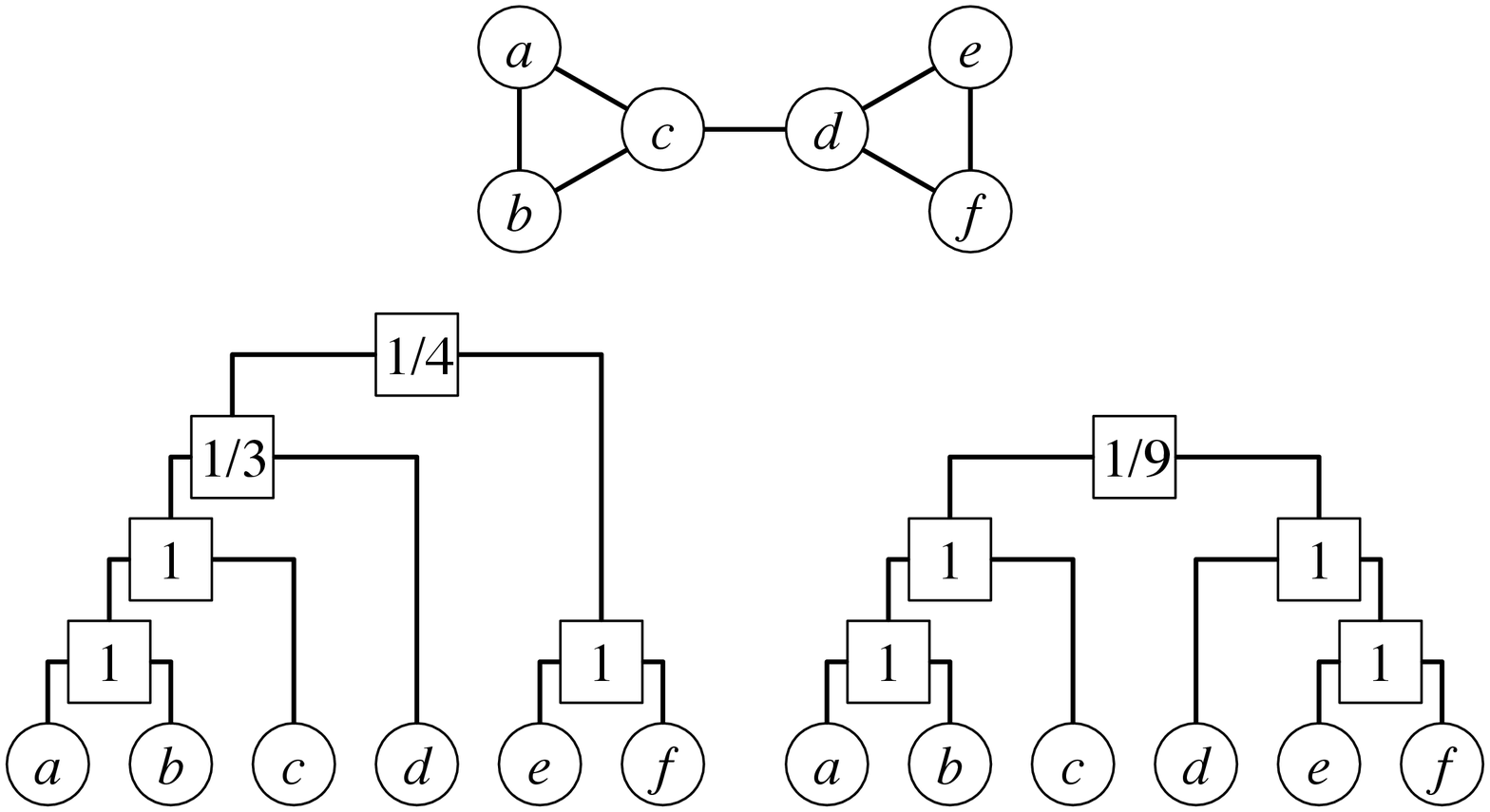}
\caption{An example network~$G$ consisting of six vertices, and the
 likelihood of two possible dendrograms.  The internal nodes~$r$ of each
 dendrogram are labeled with the maximum-likelihood probability~$\maxpr$,
 i.e.,~the fraction of potential edges between their left and right
 subtrees that exist in~$G$.  According to Eq.~\eqref{eq:like2}, the
 likelihoods of the two dendrograms are~$\like(D_1) = (1/3)(2/3)^2 \cdot
 (1/4)^2 (3/4)^6 = 0.00165\ldots$ and~$\like(D_2) = (1/9)(8/9)^8 =
 0.0433\ldots$\ \ The second dendrogram is far more likely because it
 correctly divides the network into two highly-connected subgraphs at the
 first level.}
\label{fig:small-example}
\end{center}
\end{figure}

If we fix the dendrogram~$D$, it is easy to find the
probabilities~$\{\maxpr\}$ that maximize~$\like(D,\{p_r\})$.  For each~$r$,
they are given by
\begin{equation}
\maxpr = \frac{E_r}{L_rR_r},
\label{eq:maxpr}
\end{equation}
the fraction of potential edges between the two subtrees of~$r$ that
actually appear in the graph~$G$.
The likelihood of the dendrogram evaluated at this maximum is then 
\begin{align}
\like(D) 
&= \prod_{r \in D} \left[ \maxpr^{\,\maxpr} \left(1 - \maxpr \right)^{1 -
   \maxpr} \right]^{L_r R_r}
\enspace . 
\label{eq:like2}
\end{align}
Figure~\ref{fig:small-example} shows an illustrative example, consisting of
a network with six vertices.

\begin{figure}[t]
\begin{center}
\includegraphics[scale=0.42]{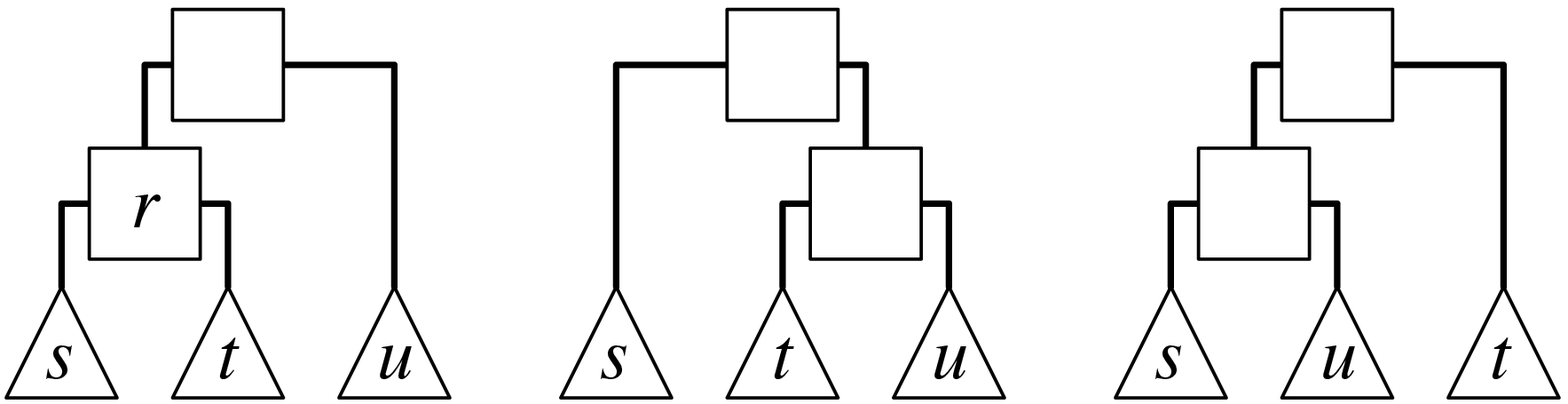}
\caption{Each internal node~$r$ of the dendrogram has three associated
 subtrees~$s$,~$t$, and~$u$, which can be placed in any of three
 configurations.  (Note that the topology of the dendrogram depends only
 on the sibling and parent relationships; the order, left to right, in
 which they are depicted is irrelevant).}
\label{fig:ergo}
\end{center}
\end{figure}

It is often convenient to work with the logarithm of the likelihood, 
\begin{align}
\log \like(D) &= - \sum_{r \in D} L_r R_r h(\maxpr),
\label{eq:loglike}
\end{align}
where~$h(p) = -p \log p - (1-p) \log (1-p)$ is the Gibbs-Shannon entropy
function.  Note that each term~$-L_r R_r h(\maxpr)$ is maximized
when~$\maxpr$ is close to~$0$ or to~$1$, i.e.,~when the entropy is
minimized.  In other words, high-likelihood dendrograms are those that
partition the vertices into groups between which connections are either
very common or very rare.

We now use a Markov chain Monte Carlo method to sample dendrograms~$D$ with
probability proportional to their likelihood~$\like(D)$.  To create the
Markov chain we need to pick a set of transitions between possible
dendrograms.  The transitions we use consist of rearrangements of subtrees
of the dendrogram as follows.  First, note that each internal node~$r$ of a
dendrogram~$D$ is associated with three subtrees: the subtrees~$s, t$
descended from its two daughters, and the subtree~$u$ descended from its
sibling.  As Figure~\ref{fig:ergo} shows, there are two ways we can reorder
these subtrees without disturbing any of their internal relationships.
Each step of our Markov chain consists first of choosing an internal
node~$r$ uniformly at random (other than the root) and then choosing
uniformly at random between the two alternate configurations of the
subtrees associated with that node and adopting that configuration.  The
result is a new dendrogram~$D'$.  It is straightforward to show that
transitions of this type are \emph{ergodic}, i.e.,~that any pair of finite
dendrograms can be connected by a finite series of such transitions.

\begin{figure*}[t]
\begin{center}
\begin{tabular}{ccc}
\includegraphics[scale=0.4]{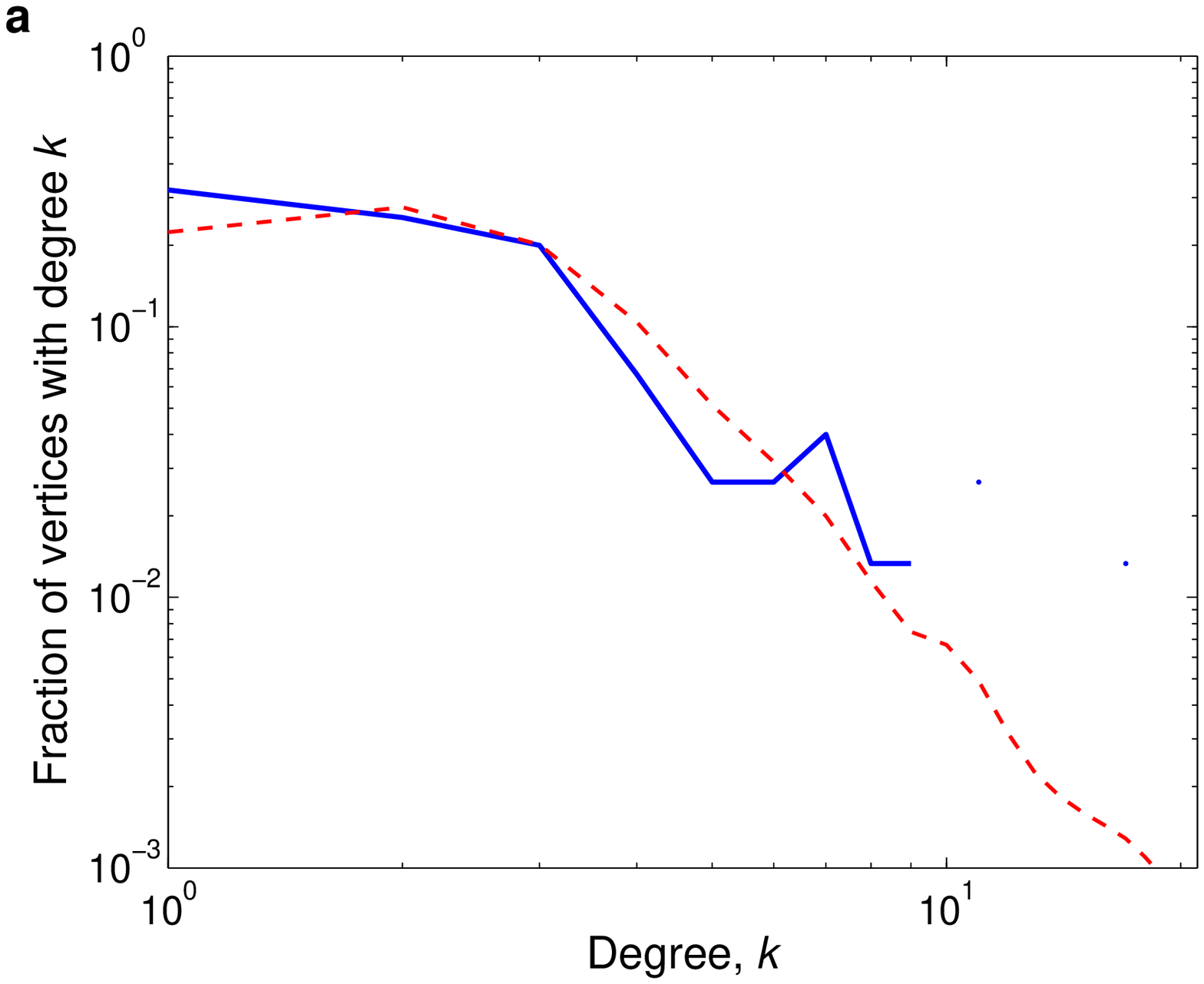} &
\includegraphics[scale=0.4]{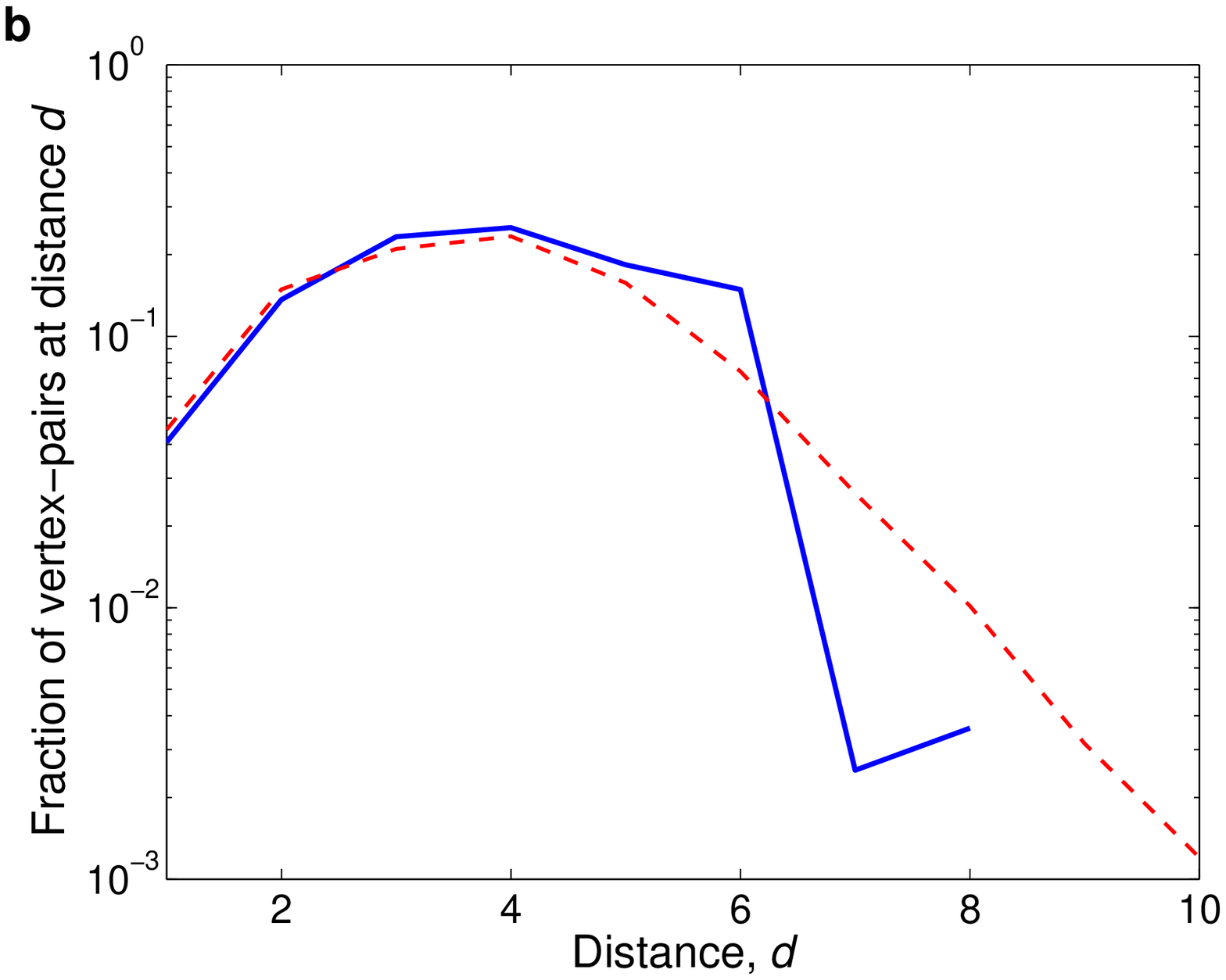} \\
\end{tabular}
\caption{Application of our hierarchical decomposition to the network of
 grassland species interactions.  \textbf{a},~Original (blue) and
 resampled (red) degree distributions.  \textbf{b},~Original and resampled
 distributions of vertex-vertex distances.}
\label{fig:valid}
\end{center}
\end{figure*}

Once we have generated our new dendrogram~$D'$ we accept or reject that
dendrogram according to the standard Metropolis--Hastings
rule.\footnote{M. E. J. Newman and G. T. Barkema, ``Monte Carlo Methods in Statistical Physics.'' Clarendon Press, Oxford (1999).}  Specifically, we accept the transition~$D
\to D'$ if~$\Delta \log \like = \log \like(D') - \log \like(D)$ is
nonnegative, so that~$D'$ is at least as likely as~$D$; otherwise we accept
the transition with probability~$\exp(\log \Delta \like) =
\like(D')/\like(D)$.  If the transition is not accepted, the dendrogram remains
the same on this step of the chain.
The Metropolis-Hastings rule ensures detailed balance and, in combination 
with the ergodicity of the transitions, guarantees a limiting probability distribution
over dendrograms that is proportional to the likelihood,~$P(D) \propto
\like(D)$.  The quantity~$\Delta \log \like$ can be calculated easily,
since the only terms in Eq.~\eqref{eq:loglike} that change from~$D$ to~$D'$
are those involving the subtrees~$s$,~$t$, and~$u$ associated with the
chosen node.  

The Markov chain appears to converge relatively quickly, with the
likelihood reaching a plateau after roughly~$O(n^2)$ steps.  This is not a
rigorous performance guarantee, however, and indeed there are mathematical
results for similar Markov chains that suggest that equilibration could
take exponential time in the worst case.\footnote{E.  Mossel and E.
 Vigoda, ``Phylogenetic MCMC Are Misleading on Mixtures of Trees.''
 \emph{Science} {\bf 309}, 2207 (2005)} Still, as our results here show,
the method seems to work quite well in practice.  The algorithm is able to
handle networks with up to a few thousand vertices in a reasonable amount
of computer time.

We find that there are typically many dendrograms with roughly equal
likelihoods, which reinforces our contention that it is important to sample
the distribution of dendrograms rather than merely focusing on the most
likely one.

\section{Resampling from the hierarchical random graph}

The procedure for resampling from the hierarchical random graph is as
follows.
\begin{enumerate}
\item Initialize the Markov chain by choosing a random starting dendrogram.

\item Run the Monte Carlo algorithm until equilibrium is reached.

\item Sample dendrograms at regular intervals thereafter from those
 generated by the Markov chain.

\item For each sampled dendrogram~$D$, create a resampled graph~$G'$
 with~$n$ vertices by placing an edge between each of the~$n(n-1)/2$
 vertex pairs~$(i,j)$ with independent probability~$p_{ij}=\maxpr$, where~$r$ is
 the lowest common ancestor of~$i$ and~$j$ in~$D$ and~$\maxpr$ is given by
 Eq.~\eqref{eq:maxpr}.  (In principle, there is nothing to prevent us from
 generating many resampled graphs from a dendrogram, but in the
 calculations described in this paper we generate only one from each
 dendrogram.)
\end{enumerate}
After generating many samples in this way, we can compute averages of
network statistics such as the degree distribution, the clustering
coefficient, the vertex-vertex distance distribution, and so forth.  Thus,
in a way similar to Bayesian model averaging,\footnote{T. Hastie, R.
 Tibshirani and J. Friedman, ``The Elements of Statistical Learning.''
 Springer, New York (2001).} we can estimate the distribution of network
statistics defined by the equilibrium ensemble of dendrograms.

\begin{figure*}[t]
\begin{center}
\includegraphics[scale=0.4]{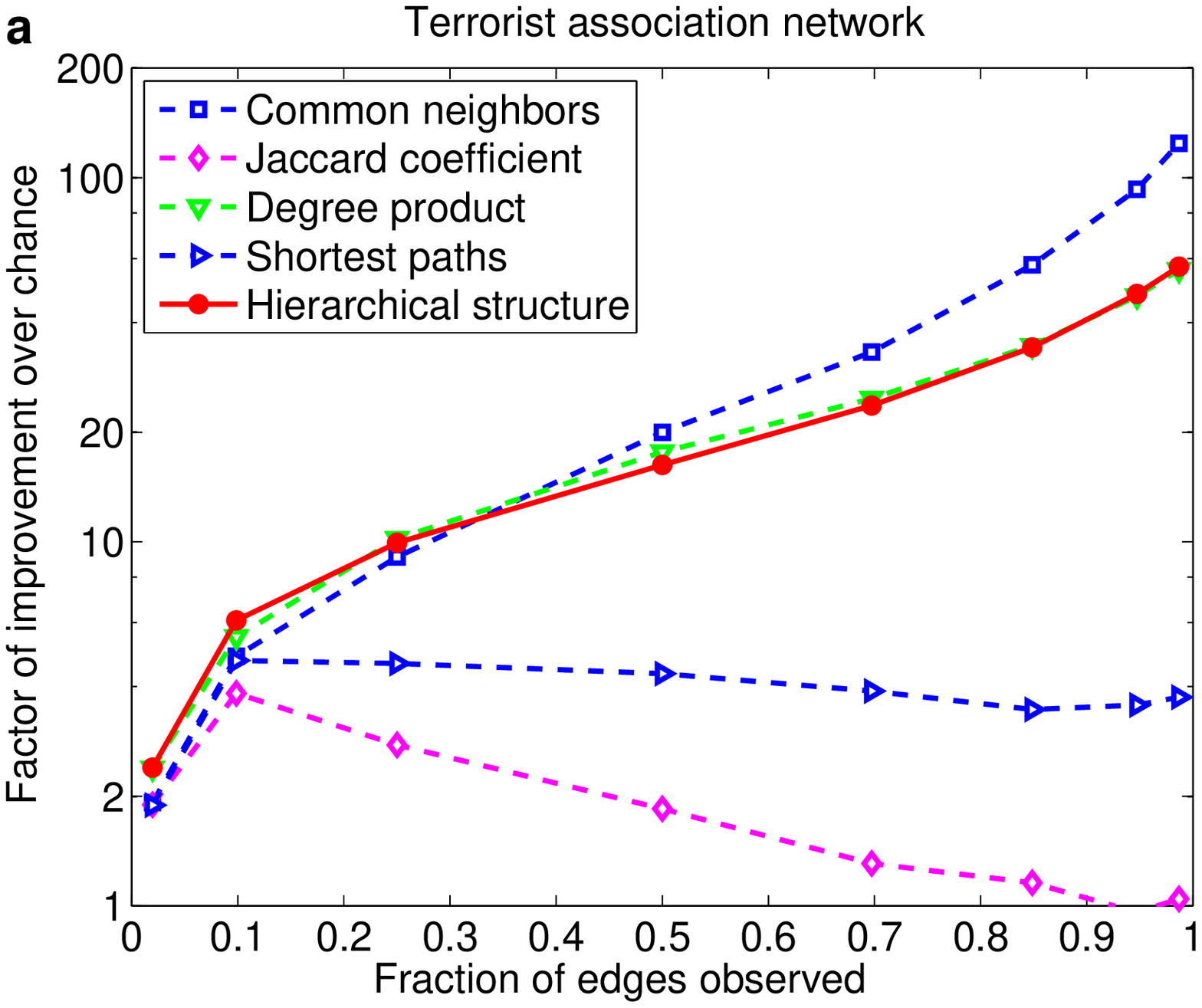} 
\includegraphics[scale=0.4]{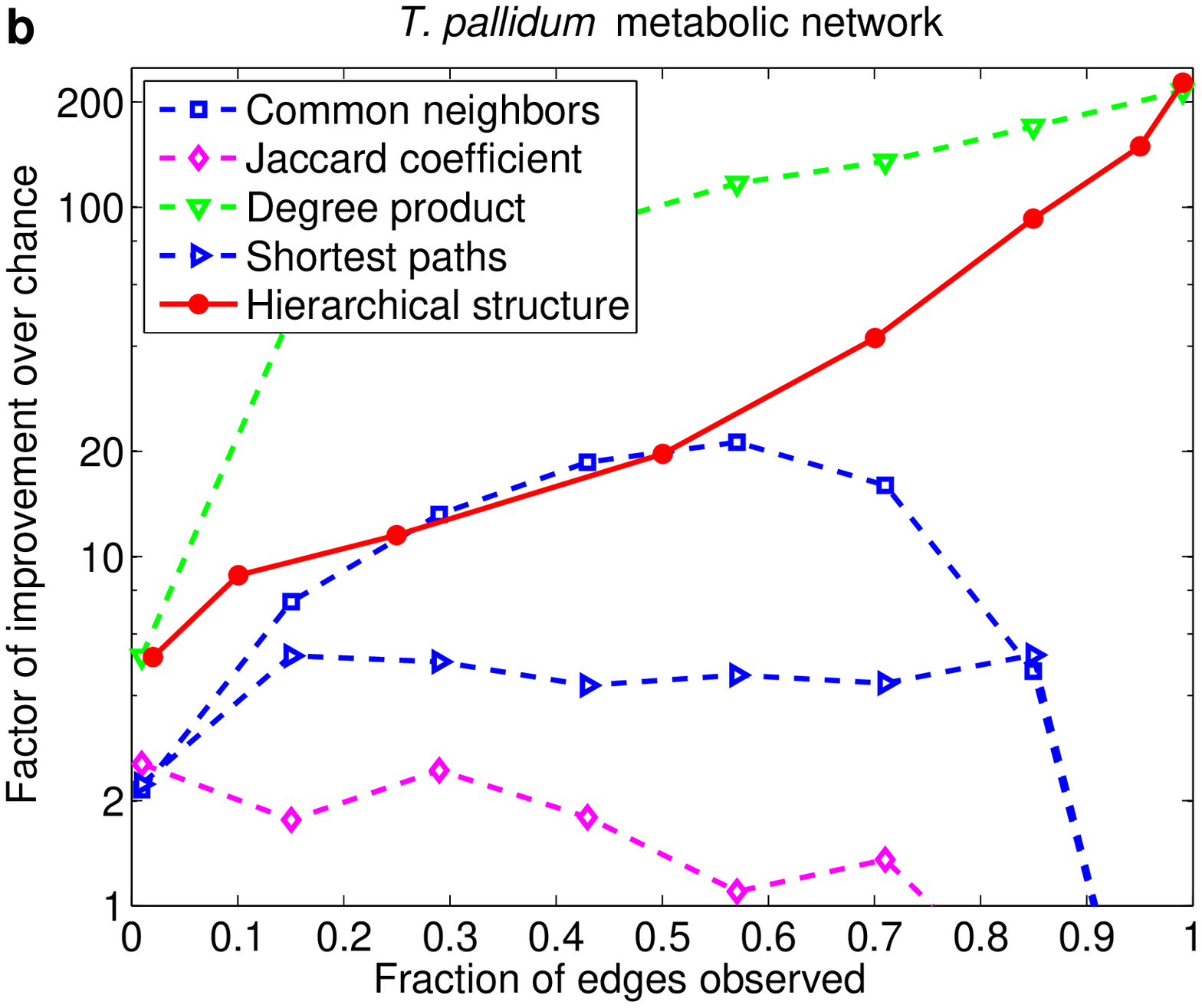} 
\includegraphics[scale=0.4]{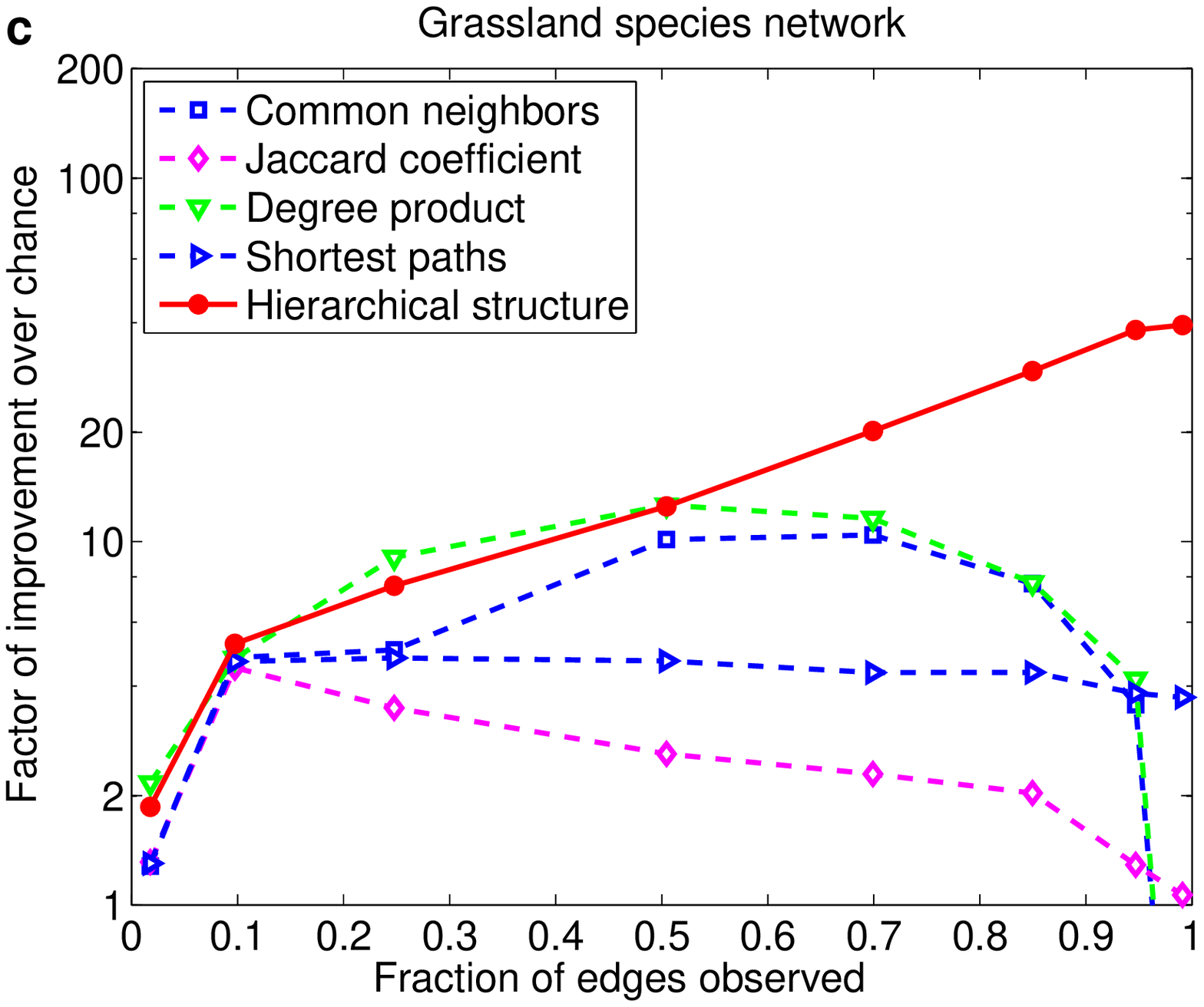} 
\caption{Further comparison of link prediction algorithms.  Data points
 represent the average ratio between the probability that the top-ranked
 pair of vertices is in fact connected and the corresponding probability
 for a randomly-chosen pair, as a function of the fraction of the
 connections known to the algorithm.  For each network, (\textbf{a},
 Terrorist associations; \textbf{b}, \emph{T. pallidum} metabolites; and
 \textbf{c}, Grassland species interactions), 
we compare our method with
 simpler methods such as guessing that two vertices are connected if they
 share common neighbors, have a high degree product, or have a short path
 between them.}
\label{fig:predictions}
\end{center}
\end{figure*}

For the construction of consensus dendrograms such as the one shown in
Fig. 2a, we found it useful to weight the most likely
dendrograms more heavily, giving them weight proportional to the square of
their likelihood, in order to extract a coherent consensus structure from
the equilibrium set of models.

\section{Predicting missing connections}

Our algorithm for using hierarchical random graphs to predict missing
connections is as follows.
\begin{enumerate}
\item Initialize the Markov chain by choosing a random starting dendrogram.

\item Run the Monte Carlo algorithm until equilibrium is reached.

\item Sample dendrograms at regular intervals thereafter from those
 generated by the Markov chain.

\item For each pair of vertices~$i,j$ for which there is not already a
 known connection, calculate the mean probability~$\langle p_{ij} \rangle$
 that they are connected by averaging over the corresponding
 probabilities~$p_{ij}$ in each of the sampled dendrograms~$D$.

\item Sort these pairs~$i,j$ in decreasing order of~$\langle p_{ij}\rangle$
 and predict that the highest-ranked ones have missing connections.
\end{enumerate}

In general, we find that the top 1\% of such predictions are highly
accurate.  However, for large networks, even the top 1\% can be an
unreasonably large number of candidates to check experimentally.  In many
contexts, researchers may want to consider using the procedure
interactively, i.e.,~predicting a small number of missing connections,
checking them experimentally, adding the results to the network, and
running the algorithm again to predict additional connections.

The alternative prediction methods we compared against, which were
previously investigated in\footnote{D. Liben-Nowell and J. Kleinberg, ``The link prediction problem for social networks.'' 
\emph{Proc. Internat. Conf. on Info. and Know. Manage.} (2003).}, consist of
giving each pair $i,j$ of vertices a score, sorting pairs in decreasing
order of their score, and predicting that those with the highest scores are
the most likely to be connected.  Several different types of scores were
investigated, defined as follows, where~$\Gamma(j)$ is the set of vertices
connected to~$j$.
\begin{enumerate}
\item Common neighbors: score$(i,j) = |\Gamma(i)\,\cap\,\Gamma(j)|$, the
 number of common neighbors of vertices~$i$ and~$j$.
\item Jaccard coefficient: score$(i,j) = |\Gamma(i)\,\cap\,\Gamma(j)| \,/\,
 |\Gamma(i)\,\cup\,\Gamma(j)|$, the fraction of all neighbors of~$i$
 and~$j$ that are neighbors of both.
\item Degree product: score$(i,j) = |\Gamma(i)|\,|\Gamma(j)|$, the product
 of the degrees of~$i$ and~$j$. 
\item Short paths: score$(i,j)$ is $1$ divided by the length of the
 shortest path through the network from~$i$ to~$j$ (or zero for vertex
 pairs that are not connected by any path).
\end{enumerate}

One way to quantify the success of a prediction method, used by previous
authors who have studied link prediction
problems${}^{\rm 7}$, is the ratio between the
probability that the top-ranked pair is connected and the probability that
a randomly chosen pair of vertices, which do not have an observed
connection between them, are connected.  Figure~\ref{fig:predictions} shows
the average value of this ratio as a function of the percentage of the
network shown to the algorithm, for each of our three networks.  Even when
fully~$50\%$ of the network is missing, our method predicts missing
connections about ten times better than chance for all three networks.  
In practical terms, this means that the amount of work required of the
experimenter to discover a new connection is reduced by a factor of~$10$,
an enormous improvement by any standard.  If a greater fraction of the
network is known, the accuracy becomes even greater, rising as high
as~$200$ times better than chance when only a few connections are missing.

We note, however, that using this ratio to judge prediction algorithms
has an important disadvantage.  Some missing connections are much easier to
predict than others: for instance, if a network has a heavy-tailed degree
distribution and we remove a randomly chosen subset of the edges, the
chances are excellent that two high-degree vertices will have a missing
connection and such a connection can be easily predicted by simple
heuristics such as those discussed above.  The AUC statistic used in the
text, by contrast, looks at an algorithm's overall ability to rank all the
missing connections over nonexistent ones, not just those that are 
easiest to predict.

Finally, we have investigated the performance of each of the prediction
algorithms on purely random (i.e., Erd\H{o}s--R\'enyi) graphs.  As
expected, no method performs better than chance in this case, since the
connections are completely independent random events and there is no
structure to discover.  We also tested each algorithm on a graph with a
power-law degree distribution generated according to the configuration
model.\footnote{M.~Molloy and B.~Reed, ``A critical point for random graphs 
with a given degree sequence.'' \textit{Random Structures and Algorithms} 
\textbf{6}, 161--179 (1995)}  In this case, guessing that high-degree vertices 
are likely to be connected performs quite well, whereas the method based on 
the hierarchical random graph performs poorly since these graphs have no 
hierarchical structure to discover.

\end{appendix}

\end{document}